\documentclass[conference]{IEEEtran}
\IEEEoverridecommandlockouts
\usepackage{cite}
\usepackage{amsmath,amssymb,amsfonts}
\usepackage{algorithm}
\usepackage{algorithmic}
\usepackage{graphicx}
\usepackage{textcomp}
\usepackage{xcolor}
\usepackage{arydshln}
\usepackage{hyperref}

\def\BibTeX{{\rm B\kern-.05em{\sc i\kern-.025em b}\kern-.08em
    T\kern-.1667em\lower.7ex\hbox{E}\kern-.125emX}}
\begin{document}

\title{Subword Semantic Hashing for Intent Classification on Small Datasets}

\author{
    Kumar Shridhar\textsuperscript{1},
    Ayushman Dash\textsuperscript{1},
    Amit Sahu\textsuperscript{1},\\
    Gustav Grund Pihlgren\textsuperscript{3},
    Pedro Alonso\textsuperscript{3},
    Vinaychandran Pondenkandath\textsuperscript{2},\\
    Gy\"{o}rgy Kov\'{a}cs\textsuperscript{3},
    Foteini Simistira\textsuperscript{2,3},
    Marcus Liwicki\textsuperscript{1,2,3}\\
    \\
    \textsuperscript{1} MindGarage, Technical University Kaiserslautern, Germany\\
    \textsuperscript{2} University of Fribourg, Switzerland\\
    \textsuperscript{3} Lule\aa \ University of Technology, Sweden\\}

\maketitle

\begin{abstract}
In this paper, we introduce the use of Semantic Hashing as embedding for the task of Intent Classification and achieve state-of-the-art performance on three frequently used benchmarks.
Intent Classification on a small dataset is a challenging task for data-hungry state-of-the-art Deep Learning based systems.
Semantic Hashing is an attempt to overcome such a challenge and learn robust text classification. Current word embedding based methods~\cite{mikolov2013distributed,mikolov2013efficient,le2014distributed} are dependent on vocabularies.
One of the major drawbacks of such methods is out-of-vocabulary terms, especially when having small training datasets and using a wider vocabulary.
This is the case in Intent Classification for chatbots, where typically small datasets are extracted from internet communication.
Two problems arise with the use of internet communication. First, such datasets miss a lot of terms in the vocabulary to use word embeddings efficiently. Second, users frequently make spelling errors.
Typically, the models for intent classification are not trained with spelling errors and it is difficult to think about ways in which users will make mistakes.
Models depending on a word vocabulary will always face such issues.
An ideal classifier should handle spelling errors inherently. With Semantic Hashing, we overcome these challenges and achieve state-of-the-art results on three datasets: Chatbot, Ask Ubuntu,  and Web Applications \cite{datasets}. Our benchmarks are available online. \footnote{\url{https://github.com/kumar-shridhar/Know-Your-Intent}}
\end{abstract}


\section{Introduction}
State-of-the-art systems in many different classification tasks have their basis in deep neural networks~\cite{Gurney:1997:INN:523781}.
This is, among other reasons, because of neural networks' ability to efficiently learn the various features present in the classes.
However, this ability also makes neural networks prone to overfitting on the training data.
A wide variety of strategies are used to prevent this, but the most reliable way to prevent overfitting is to have a large amount of training data.
This makes neural networks, deep networks in particular, unsuited for solving problems with small datasets.

With small datasets, it is often better to use less complex machine learning models. With less complex models however the feature learning of deep networks is lost. Without feature learning, the input features given to the model have a much larger impact on the model's ability to learn. In this paper, we experiment with a new feature extraction model.

One field where datasets are often small is the intent classification for industries like CRMs, Chatbots, business process automation, customer support, and so on. Intent classification is the task of giving an input, usually a text, finding the intent behind the said text. For example, the intent behind the sentence \textit{Sugar causes teeth decay} is to make you eat less sugar.

Since there are countless different intents that a text can have, labelling them is difficult and highly problem-specific. For example, political intents may have labels such as \textit{leftist} or \textit{right-wing} while questions may have intents such as \textit{where to} or \textit{how to}. Because of this, most of the real-life intent classification datasets are small (below 100 examples per class).

In this paper, we introduce an effective method for providing features to an intent classifier and we evaluate it on the Chatbot, AskUbuntu, and WebApplication corpora \cite{datasets}. A classifier trained on semantic hashing (semhash) features achieves state-of-the-art performance.

The three datasets on which the classifier has been evaluated were introduced in the paper \textit{Evaluating Natural Language Understanding Services for Conversational Question Answering Systems}  \cite{datasets} as a baseline to test Natural Languages Understanding (NLU) services.

An NLU service is a toolkit or API which can train a natural language classifier. The idea is that a user without prior knowledge of machine learning can simply provide examples of the input and the expected output of a natural language processing system, and the NLU will train that system for them. An NLU trained on intent classification data is therefore also an intent classifier.

\section{Related Work}

One of the essential parts of any deep learning based Natural Language Processing system is embeddings. Text quanta, when represented as dense learnable vectors (rather than sparse vectors), are called embeddings, which are trained with a certain predefined objective. Post training, their purpose is to be used for feature extraction and representation of text in various Machine Learning tasks. One of the most popular forms of embedding is the Word Embedding, as pioneered by models such as Word2Vec \cite{mikolov2013efficient} and GloVe \cite{pennington2014glove}.

Both of these embedding models are trained in an unsupervised fashion and are based on the distributional hypothesis: Words that occur nearby have similar contextual meaning. These word embeddings were further improved by FastText \cite{2016arXiv160701759J} with the inclusion of character n-grams. N-grams inclusion allowed better approximations of out of vocabulary words. Further, state-of-the-art was improved by Allen Institute for AI with the introduction of Deep Contextualized Word Representations (ELMo)~\cite{Peters:2018}.

Words to Sentence Embedding involves averaging of a sentence's word vectors, often referred to as the Bag of Word approach. This simple approach was further improved by the usage of Concatenated p-mean Embeddings~\cite{rueckle:2018} instead of a simple averaging. Skip-thought-vectors \cite{2015arXiv150606726K} is another approach to learning unsupervised sentence embeddings. A vocabulary expansion scheme improved their results by handling unseen words during training. The training time is very high for skip-thought-vectors. Quick thought vectors \cite{logeswaran2018an} improved the training time by replacing the decoder with a classifier by following a discriminative approximation to the generation problem.

Supervised learning approach did not seem intuitive for embeddings until InferSent~\cite{2017arXiv170502364C} used the Stanford Natural Language Inference (SNLI) Corpus~\cite{snli:emnlp2015} to train a classifier. MILA/MSR’s General Purpose Sentence Representation~\cite{2018arXiv180400079S} further extended the supervised approach by encoding multiple aspects of the same sentence. Google’s Universal Sentence Encoder~\cite{2018arXiv180311175C} uses its transformer network to train over a variety of datasets and then use the same model for a variety of tasks.

\section{Datasets}


Three different data corpora have been used for our evaluation and benchmarks: the \textit{Chatbot Corpus} (Chatbot), the \textit{Ask Ubuntu Corpus} (AskUbuntu), and the \textit{Web Applications Corpus} (WebApplication). The Chatbot corpus consists of questions written to a Telegram chatbot. The chatbot was used to answer questions regarding the public transport of Munich. The AskUbuntu and WebApplication corpora are questions and answers from StackExchange. All three corpora have predefined training and test splits. The corpora are available on GitHub under the Creative Commons CC BY-SA 3.0 license.\footnote{\url{https://github.com/sebischair/NLU-Evaluation-Corpora}}

\subsection{The Chatbot Corpus}
The Chatbot Corpus consists of two different intents \textit{(Departure Time and Find Connection)} with a total of 206 questions. The corpus also has five different entity types \textit{(StationStart, StationDest, Criterion, Vehicle, Line)} which have not been used in our benchmarks as we only focused on Intent Classification. The language of the samples present is English. However, the train station names used are in German which is evident from German vowels usage ({\"a},{\"o},{\"u},\ss). The data is further split in Train and Test datasets as shown in Table \ref{tab:data distribution Chatbot}.

\begin{table}[ht]
\caption{Data sample distribution for the Chatbot corpus}
\begin{center}
\begin{tabular}{|c | c | c|} 
\hline
Intent & Train & Test \\ [0.75ex] 
\hline
Departure Time &  43 & 35\\ 
Find Connection & 57 & 71\\
\hline
\end{tabular} 
\label{tab:data distribution Chatbot}
\end{center}
\end{table}

\subsection{The AskUbuntu Corpus}
AskUbuntu consists of five Intents \textit{(Make Update, Setup Printer, Shutdown
Computer, Software Recommendation, and None)}.
The dataset contains 190 samples that have been extracted from the AskUbuntu platform.
Only questions with the highest scores and most were extracted.
For mapping the correct Intent to these question, Amazon Mechanical Turk was used.

In addition to the questions and their labelled intent, the corpus also includes several other features: The author of the question, the URL for the page it was taken from, entities, the answer, and the author of the answer. However, none of these has been used for the benchmarks.
 
Table \ref{tab:data distribution AskUbuntu} shows the data distribution of AskUbuntu Corpus.

\begin{table}[ht]
\caption{Data sample distribution for the AskUbuntu corpus}
\begin{center}
\begin{tabular}{|c | c | c|} 
\hline
Intent & Train & Test \\ [1ex] 
\hline
Make Update &  10 & 37\\ 
 
Setup Printer &  10 & 13\\ 
     
Shutdown Computer &  13 & 14\\ 
     
Software Recommendation &  17 & 40\\ 
     
None & 3 & 5\\

\hline
\end{tabular} 
\label{tab:data distribution AskUbuntu}
\end{center}
\end{table}

\subsection{The Web Applications Corpus}
The WebApplication corpus has the same features and was prepared in the same way as AskUbuntu.
The corpus consists of 100 samples and eight Intents \textit{(Change Password, Delete Account, Download Video, Export Data, Filter Spam, Find Alternative, Sync Accounts, and None)}.
The data distribution is shown in Table \ref{tab:data distribution WebApplication}.

\begin{table}[ht]
\caption{Data sample distribution for the WebApplication corpus}
\begin{center}
\begin{tabular}{|c | c | c|} 
\hline
Intent & Train & Test \\ [1ex] 
\hline
Change Password &  2 & 6\\ 
     
Delete Account &  7 & 10\\ 
     
Download Video &  1 & 0\\
     
Export Data &  2 & 3\\
     
Filter Spam &  6 & 14\\
     
Find Alternative &  7 & 16\\ 
     
Sync Accounts &  3 & 6\\ 
     
None & 2 & 4\\

\hline
\end{tabular} 
\label{tab:data distribution WebApplication}
\end{center}
\end{table}

\section{Methodology}
\subsection{Semantic Hashing}
Our method for semantic hashing is inspired by the Deep Semantic Similarity Model~\cite{shen2014learning}.
In that work, the authors propose a way to hash tokens in an input sentence so that the model will depend on a hash value rather than on tokens.
This method also reduces hash collisions.

Our method extracts sub-word tokens (i.e. parts of words) from sentences as features. These features are then vectorized before being processed by a classifier for training or prediction. In that way, our method can be viewed as a featurizer and together with a vectorizer it can be used as an alternative to embeddings.
A description of our method is as follows:

Given an input text $T$, e.g., \textit{"I have a flying disk"}, split it into a list of words $t_i$.
The output of the split should look like, \textit{["I", "have", "a", "flying", "disk"]}.
Pass each word into a pre-hashing function $\mathcal{H}(t_i)$ to generate sub-tokens $t^j_i$, where $j$ is the index of the sub-tokens. E.g., $\mathcal{H}(have)=[\#ha, hav, ave, ve\#]$. $\mathcal{H}(t_i)$ first adds a $\#$ at the beginning and at the end of each word and then extracts trigrams from it. These trigrams are the sub-tokens $t^j_i$. This procedure is described in Algorithm \ref{alg:SSHP}.

$\mathcal{H}(t_i)$ can then be applied to the entire corpus to generate sub-tokens. These sub-tokens are then used to create a Vector Space Model (VSM). This VSM should be used to extract features for a given input text. In other words, this VSM acts as a hashing function for an input text sequence.  

\begin{algorithm}[ht]
\caption{Subword Semantic Hashing}
\begin{algorithmic}
\label{alg:SSHP}
\STATE{\textit{Texts} $\xleftarrow{}$ collection of texts}
\STATE{Create \textbf{set} \textit{sub-tokens}}
\STATE{Create \textbf{list} \textit{examples}}
\FOR{text $T$ in \textit{Texts}}
\STATE{Create \textbf{list} \textit{example}} 
\STATE{\textit{tokens} $\xleftarrow{}$ split $T$ into words.}
\FOR{token $t$ in \textit{tokens}}
\STATE{$t$ $\xleftarrow{}$ "\#" + $t$ + "\#"}
\FOR{$j$ in \textit{length}($t$)$-2$}
\STATE{Add $t$[$j$:$j+2$] to \textbf{set} \textit{sub-tokens}}
\STATE{Append $t$[$j$:$j+2$] to \textbf{list} \textit{example}}
\ENDFOR
\ENDFOR
\STATE{Append \textit{example} to \textbf{list} \textit{examples}}
\ENDFOR
\RETURN (\textit{sub-tokens}, \textit{examples})
\end{algorithmic}
\end{algorithm}

\subsection{Preprocessing and Data Augmentation}
 
 The dataset have been preprocessed by changing all letters to lower case, replacing pronouns by '-PRON-', and removing all special characters except stop characters.
 
 Dataset distribution between classes have been analyzed and less sampled classes have been oversampled by adding more augmented sentences to these classes. In the final training set, each class had an equal number of training samples for all three datasets.  
 
 The extra samples have been augmented with a dictionary-based synonym replacement of nouns and verbs chosen randomly. This helped in getting new variations in the training dataset. However, it did not take the spelling errors into account. Dictionary replacement have been done using \textit{WordNet} \cite{wordnet}.
 
 A stratified K-fold cross-validation have been performed on train dataset to obtain the training and validation split. The \textit{number of splits} value was kept as 5. 

\subsection{Vectorization}

The preprocessed text needs to be represented in the form of fixed sized numerical feature vectors to provide as an input to the classifier. The Bag of Words approach or in our case \textit{Bag of n-gram semhash tokens} approach is used to form a matrix with rows depicting the documents and columns depicting the semhash tokens occurring in the corpus of documents. The sparse term frequency based vector for the corpus consisted of frequently occurring yet uninformative semhash tokens like \textit{'\#a\#', '\#th', 'he\#'} and so on. A measure of inverse-document frequency was added to balance the occurrence of rarer yet informative semhash tokens defined as:
$$\text{idf}(t) = log \frac{1+n}{1 + df(t)} + 1$$
 where $n$ is the total number of documents in the corpus and $df(t)$ is the number of documents in the corpus that contain the token $t$. 
 
 The inverse document frequency $idf(t)$ is multiplied by the term-frequency $tf(t,d)$ as obtained above to get the final term frequency - inverse document frequency vector,  $\text{tf-idf(t,d)}$.
 
$$\text{tf-idf}(t,d) = tf(t,d) * idf(t)$$

Finally, the obtained vector is normalized by the Euclidean norm to get the final vector $v_{final}$:

$$v_{final} = \frac{v}{||v|| \textsuperscript{2}}$$

where, 

$$||v|| \textsuperscript{2} = \sqrt{v_{1}\textsuperscript{2} + v_{2}\textsuperscript{2} + ... + v_{n}\textsuperscript{2}} $$

\textbf{NOTE:} On the Chatbot, AskUbuntu, and WebApplication corpora the \textit{document} term in the above paragraph refers to one text sentence and \textit{corpus} symbolizes the entire training set. The terms \textit{document} and \textit{corpus} are used for the consistency of explanation. 

\subsection{Intent Classification}

A classifier for intents can be trained on the vectorized VSM generated by Algorithm \ref{alg:SSHP}. This classifier could be any classifier, like Support Vector Machines (SVM), Multi-Layer Perceptron (MLP), Convolutional Neural Network (CNN), etc. 


The experiments in this paper have been carried out with a number of classifiers, namely Ridge Classifier, K-Nearest Neighbors (KNN) Classifier, Multilayer Perceptron, Passive Aggressive Classifier, Random Forest Classifier, Linear Support Vector Classifier (SVC), Stochastic Gradient Descent (SGD) Classifier, Nearest Centroid, Multinomial Naive Bayes (NB), Bernoulli Naive Bayes, K-means .
The classifiers have been provided by scikit-learn library \cite{scikit-learn}.

Default parameters as provided by scikit-learn library have been used for Ridge, Passive Aggressive, Linear SVC, Nearest Centroid, Multinomial NB, Bernoulli NB, and K-means classifiers. Grid search have been used to find the best hyperparameters for MLP, Random Forest, SGD and KNN classifiers. Grid search was performed with 5-fold cross-validation on the training set and only the model with the best average validation score was applied to the test set. A prior value based on the class distribution was used for the Naive Bayes Classifiers. 
The results achieved are comparable to the state-of-the-art results for all the three corpora. To further improve the results, data augmentation was used as mentioned in the previous section.

\subsection{Evaluation}
As for performance measure, we use the micro F1-score for each dataset. For the overall performance on all datasets, we took the weighted average. For the individual datasets, this is the same as calculating the accuracy. The overall micro F1-score is calculated by finding summing the total number of true positives ($tp$), false positives ($fp$), and false negatives ($fn$) for all test sets. From that, we calculate the precision and the recall from which the F1-score is derived.
$$\text{precision}= \frac{tp}{tp+fp}$$
$$\text{recall} = \frac{tp}{tp+fn}$$
$$\text{F1-score} = 2 * \frac{\text{precision} * \text{recall}}{\text{precision} + \text{recall}}$$

To alleviate the issue of random initialization we perform 10 runs per experiment and average the performance over the runs. Note that the best performance from an individual run is even higher.


\begin{table}[t]
\caption{Micro F1 Score comparison of different NLU services with our approach}
\begin{center}
\begin{tabular}{|c|c|c|c|c|c|} 
\hline
Platform & Chatbot & AskUbuntu & WebApp & Overall & Avg. \\ [1ex] 
\hline
Botfuel  &  0.98 & 0.90 & 0.80 & 0.91 & 0.89\\ 
\hdashline     
Luis & 0.98 & 0.90 & 0.81 & 0.91 & 0.90\\ 
\hdashline      
Dialogflow  & 0.93 & 0.85 & 0.80 & 0.87 & 0.86\\
\hdashline      
Watson &  0.97 & 0.92 & 0.\textbf{83} & 0.91 & 0.91\\
\hdashline      
Rasa  &  0.98 & 0.86 & 0.74 & 0.88 & 0.86\\
\hdashline      
Snips &  0.96 & 0.83 & 0.78 & 0.89 & 0.86\\ 
\hdashline      
Recast & \textbf{0.99} & 0.86 & 0.75 & 0.89 & 0.87\\
\hdashline
TildeCNN & \textbf{0.99} & 0.92 & 0.81 & 0.92 & 0.91\\
\hline      
\textbf{Our Avg.} & \textbf{0.99} & \textbf{0.94} & \textbf{0.83} & \textbf{0.93} & \textbf{0.92}\\
\hdashline
\textbf{Our Best} & \textbf{0.996} & \textbf{0.94} & \textbf{0.85} & \textbf{0.94} & \textbf{0.93}\\
\hline
\end{tabular} 
\label{tab:f1 score comparison}
\end{center}
\end{table}

\begin{table}[t]
\caption{The best classifiers}
\begin{center}
\begin{tabular}{|c|c|c|} 
\hline
Dataset & Best classifier & Accuracy\\ \hline
All Averaged & Ridge Classifier & 0.92\\ 
             & Linear Support Vector Classifier & 0.92\\ \hdashline
Chatbot & Passive Aggressive Classifier & 0.996\\ \hdashline
AskUbuntu & Ridge Classifier & 0.94\\ 
          & Linear Support Vector Classifier & 0.94\\ \hdashline
WebApp & Random Forest Classifier & 0.85\\
\hline
\end{tabular} 
\label{tab:best classifier}
\end{center}
\end{table}

\section{Results and Analysis}
The performance of our method is evaluated on the test dataset on all three corpora. Specifically, we evaluate accuracy per dataset, the mean accuracy of all three datasets, and the overall micro-F1 score on all three datasets.

The performance is compared against the results on various NLU services and open source NLU platforms in the market: Botfuel, Dialogflow, Luis, Watson, Rasa, Recast, and Snips. In addition, we compare the performance against a recently published classifier dubbed TildeCNN \cite{10.1007/978-3-319-99344-7_3}.
Results for Dialogflow, Watson, Rasa, and Luis comes from \cite{datasets} which have benchmarked the initial results. \cite{DBLP:journals/corr/abs-1805-10190} reproduced the results for Watson, Rasa, Dialogflow and Luis and compared it with Snips platform.
Finally, the result comparison table was extended by \cite{botfuel_blog} with the inclusion of Recast, a bot building platform and Botfuel, an NLP classification service. 

Table~\ref{tab:f1 score comparison} shows the comparisons of the evaluation. 
Our results are presented in two rows.
\textbf{Our Avg.} shows the performance of the best single classifier, i.e. the classifier with achieves the best average performance (average of all three datasets).
\textbf{Our Best} shows the performance of the best classifier for each dataset (still averaged over 10 runs).

The best classifiers for each dataset appear in table~\ref{tab:best classifier}. The accuracy of all classifiers on Chatbot, AskUbuntu, and WebApplication can be found in Tables \ref{tab:chatbot classifiers}, \ref{tab:askubuntu classifiers}, and \ref{tab:webapplication classifiers} respectively.

A noteworthy result of the tests is that all the tested classifiers have extremely low (less than $10^{-3}$) variance between runs.

Another important result is that several classifiers managed to perfectly predict the classes of the Chatbot test set during some of the 10 runs. The best classifier on Chatbot, Passive Aggressive Classifier, managed this feat 7 out of the 10 runs.

We achieved state-of-the-art results on all three corpora and outperformed the previous state-of-the-art on AskUbuntu corpus as well as the average and overall performance on all three datasets.


The preprocessing, featurizing, and training of the data is in the order of seconds. Even more impressive is the inference time which is in the order of milliseconds. The training and test times for the different classifiers on the three datasets can be found in Tables \ref{tab:chatbot classifiers}, \ref{tab:askubuntu classifiers}, and \ref{tab:webapplication classifiers}. The times in these tables are the training time on the entire training set and the test time on the entire test set. Preprocessing and featurizing times for each dataset can be found in table \ref{tab:preprocess time}. Preprocessing refers to the process of loading, oversampling, and making synonyms replacements on the dataset. Featurizing refers to extracting the trigrams according to Algorithm \ref{alg:SSHP} and vectorizing. Unfortunately, there is no data available for training and testing time for other NLU services and hence no comparison can be made.

\section{Discussion and Future Work}

The micro F1-score of the three big companies (Dialogflow, Luis, and Watson) vary a lot on the three corpora, showing a lot of difference in the approach used by these companies.
One interesting thing to note is that the results are comparable to the start-ups: Recast, and Botfuel, as well as, the open source platforms, like Snips and Rasa. This shows that this field of AI is rather new for everyone and there is no clear domination by the giants in the field.

From our results, we can tell that our method is accurate, versatile, stable, and fast. Our method performs well on all three corpora and achieves the best results on average, which shows good potential for the method. Additionally, the method can achieve high accuracy with a wide variety of classifiers proving its versatility. The extremely low variance of performance across all the tested classifiers suggests that the method is stable. The quick training of the model allows the user to select the classifier that best suits the problem at hand. The whole solution if integrated into a conversational service will act in real time.

In future work, semhash needs to be tested and to be benchmarked on more datasets to assess the domain independence of the method. Furthermore, semhash should be compared with other feature extraction methods to determine where semhash is the preferred choice.

\begin{table}[htbp]
\caption{Performance of all classifiers on Chatbot}
\begin{center}
\begin{tabular}{|l|c|c|c|} 
\hline
Classifier & Avg. Acc & Train time & Test time \\ \hline 
Ridge Classifier    & 0.99 & 0.03 s     & 2.5 ms \\ \hdashline 
KNN Classifier      & 0.94 & 1.36 s     & 66.4 ms\\ \hdashline 
MLP                 & 0.99 & 37.21 s    & 5.5 ms \\ \hdashline 
Passive Aggressive  & 0.996 & 0.06 s     & 3.0 ms\\ \hdashline 
Random Forest       & 0.95 & 2.43 s     & 7.2 ms\\ \hdashline 
Linear SVC          & 0.99 & 0.01 s     & 1.6 ms\\ \hdashline 
SGD Classifier      & 0.99 & 0.10 s     & 3.3 ms\\ \hdashline 
Nearest Centroid    & 0.97 & 0.00 s     & 3.5 ms\\ \hdashline 
Multinomial NB      & 0.99 & 0.01 s     & 4.2 ms\\ \hdashline 
Bernoulli NB        & 0.99 & 0.01 s     & 8.6 ms\\ \hdashline 
K-means             & 0.04 & 0.24 s     & 5.2 ms\\ \hline 
\end{tabular} 
\label{tab:chatbot classifiers}
\end{center}
\end{table}

\begin{table}[htbp]
\caption{Performance of all classifiers on AskUbuntu}
\begin{center}
\begin{tabular}{|l|c|c|c|} 
\hline
Classifier & Avg. Acc & Train time & Test time \\ \hline 
Ridge Classifier    & 0.94 & 0.06 s & 3.7 ms\\ \hline 
KNN Classifier      & 0.85 & 1.09 s & 69.0 ms\\ \hdashline 
MLP                 & 0.89 & 37.88 s & 19.8 ms\\ \hdashline 
Passive Aggressive  & 0.93 & 0.26 s & 2.8 ms\\ \hdashline 
Random Forest       & 0.91 & 2.50 s & 7.0 ms\\ \hdashline 
Linear SVC          & 0.94 & 0.03 s & 2.8 ms\\ \hdashline 
SGD Classifier      & 0.91 & 0.45 s & 2.6 ms\\ \hdashline 
Nearest Centroid    & 0.91 & 0.00 s & 7.4 ms\\ \hdashline 
Multinomial NB      & 0.90 & 0.01 s & 4.5 ms\\ \hdashline 
Bernoulli NB        & 0.91 & 0.01 s & 8.3 ms\\ \hdashline 
K-means             & 0.10 & 0.22 s & 7.4 ms\\ \hline 
\end{tabular} 
\label{tab:askubuntu classifiers}
\end{center}
\end{table}

\begin{table}[htbp]
\caption{Performance of all classifiers on WebApplication}
\begin{center}
\begin{tabular}{|l|c|c|c|} 
\hline
Classifier & Avg. Acc & Train time & Test time \\ \hline 
Ridge Classifier    & 0.83 & 0.11 s & 2.5 ms\\ \hdashline 
KNN Classifier      & 0.69 & 0.32 s & 16.2 ms\\ \hdashline 
MLP                 & 0.77 & 29.23 s & 4.8 ms\\ \hdashline 
Passive Aggressive  & 0.81 & 0.17 s & 5.0 ms\\ \hdashline 
Random Forest       & 0.85 & 2.26 s & 5.6 ms\\ \hdashline 
Linear SVC          & 0.83 & 0.01 s & 1.4 ms\\ \hdashline 
SGD Classifier      & 0.83 & 0.09 s & 3.6 ms\\ \hdashline 
Nearest Centroid    & 0.79 & 0.00 s & 2.0 ms\\ \hdashline 
Multinomial NB      & 0.76 & 0.00 s & 4.6 ms\\ \hdashline 
Bernoulli NB        & 0.80 & 0.00 s & 6.9 ms\\ \hdashline 
K-means             & 0.02 & 0.07 s & 2.3 ms\\ \hline 
\end{tabular} 
\label{tab:webapplication classifiers}
\end{center}
\end{table}

\begin{table}[htbp]
\caption{Time of the data preparation for all datasets}
\begin{center}
\begin{tabular}{|l|c|c|c|} 
\hline
Dataset & Preprocessing & Featurizing & Total \\ \hline
Chatbot & 3.55 s & 5.09s & 8.64 s \\ \hdashline
AskUbuntu & 3.19 s & 4.03 s & 7.22 s \\ \hdashline
WebApplication & 1.14 s & 1.87 s & 3.02 s \\ \hline
\end{tabular} 
\label{tab:preprocess time}
\end{center}
\end{table}

\bibliographystyle{plain}
\bibliography{references}

\begin{thebibliography}{10}

\bibitem{10.1007/978-3-319-99344-7_3}
Kaspars Balodis and Daiga Deksne.
\newblock Intent detection system based on word embeddings.
\newblock In Gennady Agre, Josef van Genabith, and Thierry Declerck, editors,
  {\em Artificial Intelligence: Methodology, Systems, and Applications}, pages
  25--35, Cham, 2018. Springer International Publishing.

\bibitem{snli:emnlp2015}
Samuel~R. Bowman, Gabor Angeli, Christopher Potts, and Christopher~D. Manning.
\newblock A large annotated corpus for learning natural language inference.
\newblock In {\em Proceedings of the 2015 Conference on Empirical Methods in
  Natural Language Processing (EMNLP)}. Association for Computational
  Linguistics, 2015.

\bibitem{datasets}
Daniel Braun, Adrian Hernandez-Mendez, Florian Matthes, and Manfred Langen.
\newblock Evaluating natural language understanding services for conversational
  question answering systems.
\newblock In {\em Proceedings of the 18th Annual SIGdial Meeting on Discourse
  and Dialogue}, pages 174--185, 2017.

\bibitem{2018arXiv180311175C}
D.~{Cer}, Y.~{Yang}, S.-y. {Kong}, N.~{Hua}, N.~{Limtiaco}, R.~{St.~John},
  N.~{Constant}, M.~{Guajardo-Cespedes}, S.~{Yuan}, C.~{Tar}, Y.-H. {Sung},
  B.~{Strope}, and R.~{Kurzweil}.
\newblock {Universal Sentence Encoder}.
\newblock {\em ArXiv e-prints}, March 2018.

\bibitem{2017arXiv170502364C}
A.~{Conneau}, D.~{Kiela}, H.~{Schwenk}, L.~{Barrault}, and A.~{Bordes}.
\newblock {Supervised Learning of Universal Sentence Representations from
  Natural Language Inference Data}.
\newblock {\em ArXiv e-prints}, May 2017.

\bibitem{DBLP:journals/corr/abs-1805-10190}
Alice Coucke, Alaa Saade, Adrien Ball, Th{\'{e}}odore Bluche, Alexandre
  Caulier, David Leroy, Cl{\'{e}}ment Doumouro, Thibault Gisselbrecht,
  Francesco Caltagirone, Thibaut Lavril, Ma{\"{e}}l Primet, and Joseph Dureau.
\newblock Snips voice platform: an embedded spoken language understanding
  system for private-by-design voice interfaces.
\newblock {\em CoRR}, abs/1805.10190, 2018.

\bibitem{wordnet}
Christiane Fellbaum.
\newblock {\em WordNet: an electronic lexical database}.
\newblock MIT Press, 2000.

\bibitem{Gurney:1997:INN:523781}
Kevin Gurney.
\newblock {\em An Introduction to Neural Networks}.
\newblock Taylor \& Francis, Inc., Bristol, PA, USA, 1997.

\bibitem{2016arXiv160701759J}
A.~{Joulin}, E.~{Grave}, P.~{Bojanowski}, and T.~{Mikolov}.
\newblock {Bag of Tricks for Efficient Text Classification}.
\newblock {\em ArXiv e-prints}, July 2016.

\bibitem{2015arXiv150606726K}
R.~{Kiros}, Y.~{Zhu}, R.~{Salakhutdinov}, R.~S. {Zemel}, A.~{Torralba},
  R.~{Urtasun}, and S.~{Fidler}.
\newblock {Skip-Thought Vectors}.
\newblock {\em ArXiv e-prints}, June 2015.

\bibitem{le2014distributed}
Quoc Le and Tomas Mikolov.
\newblock Distributed representations of sentences and documents.
\newblock In {\em International Conference on Machine Learning}, pages
  1188--1196, 2014.

\bibitem{logeswaran2018an}
Lajanugen Logeswaran and Honglak Lee.
\newblock An efficient framework for learning sentence representations.
\newblock In {\em International Conference on Learning Representations}, 2018.

\bibitem{mikolov2013efficient}
Tomas Mikolov, Kai Chen, Greg Corrado, and Jeffrey Dean.
\newblock Efficient estimation of word representations in vector space.
\newblock {\em arXiv preprint arXiv:1301.3781}, 2013.

\bibitem{mikolov2013distributed}
Tomas Mikolov, Ilya Sutskever, Kai Chen, Greg~S Corrado, and Jeff Dean.
\newblock Distributed representations of words and phrases and their
  compositionality.
\newblock In {\em Advances in neural information processing systems}, pages
  3111--3119, 2013.

\bibitem{botfuel_blog}
Trong~Canh Nguyen.
\newblock Benchmarking intent classification services - june 2018.
\newblock
  \url{https://medium.com/botfuel/benchmarking-intent-classification-services-june-2018-eb8684a1e55f},
  Jun 2018.

\bibitem{scikit-learn}
F.~Pedregosa, G.~Varoquaux, A.~Gramfort, V.~Michel, B.~Thirion, O.~Grisel,
  M.~Blondel, P.~Prettenhofer, R.~Weiss, V.~Dubourg, J.~Vanderplas, A.~Passos,
  D.~Cournapeau, M.~Brucher, M.~Perrot, and E.~Duchesnay.
\newblock Scikit-learn: Machine learning in {P}ython.
\newblock {\em Journal of Machine Learning Research}, 12:2825--2830, 2011.

\bibitem{pennington2014glove}
Jeffrey Pennington, Richard Socher, and Christopher~D Manning.
\newblock Glove: Global vectors for word representation.
\newblock In {\em EMNLP}, volume~14, pages 1532--1543, 2014.

\bibitem{Peters:2018}
Matthew~E. Peters, Mark Neumann, Mohit Iyyer, Matt Gardner, Christopher Clark,
  Kenton Lee, and Luke Zettlemoyer.
\newblock Deep contextualized word representations.
\newblock In {\em Proc. of NAACL}, 2018.

\bibitem{rueckle:2018}
Andreas R{\"u}ckl{\'e}, Steffen Eger, Maxime Peyrard, and Iryna Gurevych.
\newblock Concatenated power mean embeddings as universal cross-lingual
  sentence representations.
\newblock {\em arXiv}, 2018.

\bibitem{shen2014learning}
Yelong Shen, Xiaodong He, Jianfeng Gao, Li~Deng, and Gr{\'e}goire Mesnil.
\newblock Learning semantic representations using convolutional neural networks
  for web search.
\newblock In {\em Proceedings of the 23rd International Conference on World
  Wide Web}, pages 373--374. ACM, 2014.

\bibitem{2018arXiv180400079S}
S.~{Subramanian}, A.~{Trischler}, Y.~{Bengio}, and C.~J {Pal}.
\newblock {Learning General Purpose Distributed Sentence Representations via
  Large Scale Multi-task Learning}.
\newblock {\em ArXiv e-prints}, March 2018.

\end{thebibliography}

\end{document}